\pdfoutput=1
\documentclass[letterpaper, 10 pt, conference]{ieeeconf}

\IEEEoverridecommandlockouts

\usepackage{graphicx}
\usepackage{amsmath}
\usepackage{amssymb}
\usepackage{cite}
\usepackage{xcolor}
\usepackage{booktabs}
\usepackage[hidelinks]{hyperref}
\usepackage[nameinlink,noabbrev]{cleveref}
\usepackage{mathtools}
\let\labelindent\relax
\usepackage{enumitem}
\usepackage{placeins}
\usepackage{array}


\newcommand{\methodname}{{\textsc{\textbf{Drift}}}}
\newcommand{\methodnameFull}{{\textit{\textbf{D}irect \textbf{R}einforcement Learning from \textbf{I}mages \textbf{F}or \textbf{T}ransfer}}}

\newcommand{\projectwebsite}{\par\noindent Project website: \href{https://jmarangola.github.io/one-phase-visual-rl/}{\mbox{jmarangola.github.io/one-phase-visual-rl}}}

\title{\LARGE \bf
\textsc{One Phase Is Enough}: Sim-to-Real Transfer with \\Direct Visual Reinforcement Learning
}

\author{%
John Marangola$^{*\,1,\,2}$ \quad
Antonia Bronars$^{1,\,2}$ \quad
Younghyo Park$^{1,\,2}$ \\[0.2em]
Stephen Andrews$^{1,\,2}$ \quad
Kushal Kedia$^{1,\,2\,}$ \quad
Pulkit Agrawal$^{1,\,2}$ \\[0.6em]
{\small $^{1}$Massachusetts Institute of Technology
\quad $^{2}$Improbable AI Lab}%
\thanks{$^{*}$Corresponding author. Email: \texttt{jmgola@mit.edu}}%
}

\begin{document}

\maketitle
\thispagestyle{empty}
\pagestyle{empty}

\begingroup
\setlength{\parskip}{0pt}
\begin{abstract}
Sim-to-real reinforcement learning (RL) has enabled impressive robotic behaviors, yet deploying policies directly from vision remains challenging. A common approach first trains a teacher policy with RL using privileged state observations, then distills its behavior into a vision-based student via imitation. However, the resulting student can substantially underperform its teacher. Directly training visual policies with RL offers a promising alternative, but learning task-relevant visual representations from scratch can be sample-inefficient, making one-phase visual RL intractable.

We propose \methodname, a simple recipe for efficient one-phase visual RL using image observations available at test time. Our key insight is to use frozen, pretrained foundation vision encoders, allowing RL to focus on learning control from existing visual representations. Freezing the encoder avoids costly backpropagation through the vision backbone, while its pretrained representations enable real-world transfer from fast, rasterized simulation renders. Under matched architectures, observations, and compute budgets, \methodname{} outperforms two-phase distillation and achieves a 58.8\% relative improvement in average real-world success rate across three manipulation tasks. Controlled simulation evaluations demonstrate that \methodname{}'s relative advantage grows from roughly 9\% in-distribution to over 61\% under out-of-distribution shifts, suggesting that stronger generalization contributes to superior real-world transfer.\projectwebsite

\end{abstract}
\endgroup
\section{Introduction}

Sim-to-real reinforcement learning (RL) provides a scalable alternative to real-world data collection by enabling massive amounts of parallel trial-and-error in simulation. This paradigm has pushed robot hardware toward its performance limits in locomotion~\cite{hwangbo2019learning,tan2018sim,margolis2024rapid,radosavovic2024real} and dexterous manipulation~\cite{akkaya2019solving,tang2023industreal, kedia2026simtoolreal}. However, these successes largely rely on compact state observations, such as object poses, which are estimated through heavily engineered perception pipelines.

\begin{figure}[!t]
    \vspace*{3.5bp}
    \centering
    \includegraphics[width=\linewidth]{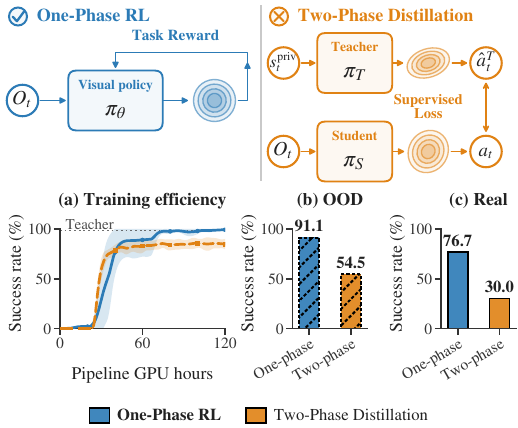}
    \caption{\textbf{One-phase visual RL closes the realizability gap, generalizes better outside of the training distribution and exhibits superior transfers to the real-world over two-phase approaches.}
    Policy learning paradigm schematics above; Push-T results below:
    (a) full-pipeline L40S-equivalent GPU hours, with a 2-hour EMA half-life per seed;
    (b) Out of distribution evaluation (OOD) in simulation; (c) real-world transfer.
    Curves show three-seed means $\pm$ one standard deviation; OOD bars show three-seed means; real-world bars show separately evaluated policies (30 trials each).
    Dotted line: privileged teacher performance.}
    \label{fig:paper_teaser}
\end{figure}
\begin{figure*}[t!]
    \centering
    \includegraphics[width=\textwidth]{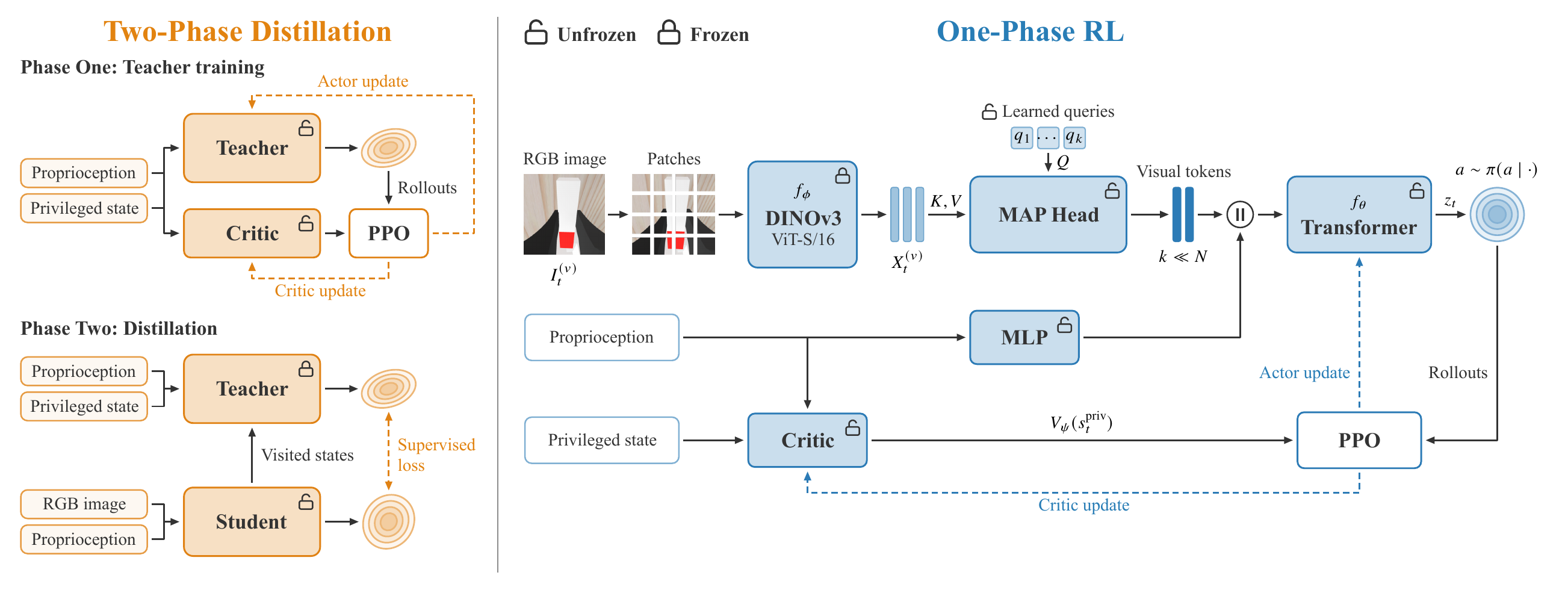}
    \caption{\textbf{DRIFT learns visual policies directly from task rewards.} Simulation provides privileged state information, such as object poses and velocities, that is not directly available at deployment. Two-phase approaches use this information to train a teacher, then train a visual student to imitate its actions (left). We instead optimize the visual policy directly with RL, using a frozen pretrained vision encoder to make training efficient (right).
    }
    \label{fig:method}
\end{figure*}

To deploy policies directly from visual observations, a common approach is to distill privileged state-based teacher policies into image-based students~\cite{chen2023visual}. The teacher is first trained with RL using privileged simulator state, after which its behavior is transferred to the visual student via imitation. However, this two-phase process introduces two sources of performance degradation. First, the privileged teacher may rely on information unavailable to the student, creating a realizability gap that prevents faithful policy distillation~\cite{ankile2025imitation}. Second, although imitation-trained students can achieve high success rates in simulation, they may generalize poorly across the sim-to-real gap~\cite{shenfeld2026rl}. Consequently, distilled visual policies can substantially underperform their state-based RL teachers upon real-world deployment~\cite{yin2026emergent}.

One-phase visual RL offers the appealing alternative of directly optimizing the policy with image observations available at test time. However, this couples the two difficult problems of learning task-relevant visual representations and control policies, with training the vision encoder from scratch typically dominating the computational cost~\cite{chen2021improving}. We address this challenge by building on pretrained foundation vision encoders~\cite{simeoni2025dinov3}, allowing RL to focus on learning control from existing visual representations. Freezing the encoder eliminates backpropagation through the vision backbone, while its pretrained representations enable policies trained on fast, rasterized simulation renders to transfer to real-world images. Together, these design choices enable one-phase visual RL to outperform two-phase distillation under matched architectures, observations, and compute budgets.

Beyond simulation performance, we investigate how one-phase visual RL and two-phase distillation differ in real-world transfer and generalization capabilities. \textbf{\emph{Our key finding is that even when the two approaches achieve comparable success in simulation, one-phase visual RL transfers substantially better to the real world, achieving a 58.8\% relative improvement in average success rate over two-phase distillation across three manipulation tasks.}} To investigate this gap, we evaluate both approaches under controlled in-distribution (ID) and out-of-distribution (OOD) conditions in simulation, varying dynamics, visual appearance, and initial states. One-phase visual RL achieves roughly 9\% higher average ID success, consistent with avoiding the realizability gap introduced by distillation from privileged state-based teachers. This relative advantage grows to over 61\% under OOD conditions, suggesting that the stronger real-world transfer reflects a broader advantage in generalization beyond the training distribution. Our contributions are:

\begin{itemize}[leftmargin=*, itemsep=0pt,
    topsep=0pt, parsep=0pt, partopsep=0pt]
    \item We introduce \methodname: \methodnameFull, a simple recipe for efficient one-phase visual RL using frozen foundation vision encoders.
\item We demonstrate a generalization gap between one-phase visual RL and two-phase distillation through controlled simulation evaluations, showing that \methodname\ achieves a 61\% relative improvement in OOD settings.
\item We validate \methodname{} across three real-world manipulation tasks, achieving a 58.8\% relative improvement in average success rate over two-phase distillation and show generalization to novel object instances unseen during training.
\end{itemize}
\section{Related Work}
\textbf{Teacher-Student Distillation.}
Privileged-policy distillation has enabled sim-to-real transfer for dexterous grasping, multi-fingered manipulation, and humanoid loco-manipulation~\cite{singh2024dextrah,bauza2025demostart,he2025viral}. However, strong teacher performance does not guarantee equally capable students~\cite{ankile2025imitation}. A typical limitation is information asymmetry (realizability gap), in which teacher actions depend on state that the student cannot infer from its observations~\cite{kim2025distilling}. To address this issue, Messikommer et al.~\cite{messikommer2025studentinformedteachertraining} jointly train teacher and student policies, adding a teacher--student action-disagreement penalty to the teacher's reward and a supervised loss that aligns the teacher with the student, encouraging behaviors that are realizable from the student's limited observations. Singh et al.~\cite{singh2025end} train depth-based teachers with RL before distilling them into RGB policies, improving performance over distillation from state-based teachers. Our approach, instead, directly trains the deployed RGB policy with RL. Furthermore, we examine the robustness limitations of distillation beyond the well-established realizability gap.

\textbf{Efficient Visual RL and Frozen Representations.}
Direct visual RL has become increasingly tractable with advances in GPU-parallelized simulation and rendering~\cite{zakka2025mujoco,tao2025demonstrating}.
Prior work improves training efficiency through low-resolution images and lightweight encoders~\cite{almuzairee2026squint}, encoder freezing and feature caching~\cite{chen2021improving}, and demonstrations that initialize policies and guide RL updates~\cite{shah2021rrl}.
Frozen pretrained encoders provide another way to simplify visual RL by removing the need to learn visual features from task rewards~\cite{yuan2022pre}.
Most closely related, PoliFormer~\cite{zeng2024poliformer} combines frozen DINOv2 features with on-policy RL for zero-shot sim-to-real navigation.
We build on this approach for contact-rich manipulation, where policies must extract precise spatial information from visual observations.
Our recipe combines learned attention pooling over frozen patch features with fast rasterized rendering, reducing the computational burden of visual control while preserving access to spatially detailed representations.

\textbf{Generalization in RL vs. Supervised Learning.}
Studies of foundation model post-training find that RL can improve OOD generalization and preserve prior capabilities better than supervised fine-tuning~\cite{chu2025sftmemorizesrlgeneralizes,shenfeld2026rl}.
In robotics, Liu et al.~\cite{liu2025what} similarly find that RL fine-tuning improves semantic generalization and execution robustness over supervised fine-tuning of VLAs.
Whereas these studies examine adaptation of pretrained models, we compare direct visual RL with on-policy teacher distillation under matched architectures, observations, and compute budgets.
\section{Preliminaries}
At time $t$, the robot receives an observation $o_t=(I_t,p_t)$
comprising RGB images $I_t$ and proprioception $p_t$, and selects
a continuous action $a_t$. A reward $r_t$ specifies the desired
task behavior. We compare two ways of learning a visual policy
from simulation: directly optimizing its task return with RL,
and training it to imitate a privileged RL teacher.

\subsection{Direct Visual Reinforcement Learning}
Direct visual RL (Fig. \ref{fig:method}, right) learns a policy $\pi_\theta(a_t\mid o_t)$ by
maximizing its expected discounted return,
\begin{equation}
    J(\theta) =
    \mathbb{E}_{\tau\sim p_{\pi_\theta}}
    \left[\sum_{t=0}^{T-1}\gamma^t r_t\right],
    \qquad \max_\theta J(\theta),
    \label{eq:rl_objective}
\end{equation}
where $p_{\pi_\theta}$ is the trajectory distribution induced by
the policy in simulation, $T$ is the episode horizon, and
$\gamma\in[0,1]$ is the discount factor. The learning objective therefore evaluates actions through their consequences for task
return.

\subsection{Two-Phase Policy Distillation}
Two-phase training first learns a teacher
$\pi_T(a_t\mid s_t^{\mathrm{priv}})$ using the same
expected-return objective, but with privileged simulator state
such as object poses and velocities as observations. The second phase learns a
visual student $\pi_{S,\theta}(a_t\mid o_t)$ by mimicking the teacher's actions. In the DAgger-style~\cite{ross2011reductionimitationlearningstructured}
distillation considered here (Fig. \ref{fig:method}, left),
the current student generates trajectories and the frozen teacher
labels the visited states. The student policy is updated via:
\begin{equation}
    \theta^{i+1} = \arg\max_{\theta}\;
    \mathbb{E}_{\tau\sim p_{\pi_{S,\theta^i}}}
    \left[
    \frac{1}{T}\sum_{t=0}^{T-1}
    \log \pi_{S,\theta}(a_t^T\mid o_t)
    \right]
    \label{eq:dagger_objective}
\end{equation}
where $a_t^T=\mu_T(s_t^{\mathrm{priv}})$ is the teacher's
mean action, and $\theta^i$ denotes the student parameters
used to collect trajectories. Although data collection is
on-policy, the student is optimized to reproduce teacher
actions rather than maximize its own task return.

\section{Method}
We propose \methodname{}, a tractable recipe for one-phase visual RL and zero-shot sim-to-real transfer (Fig. ~\ref{fig:method}). We call the approach
''one phase`` because it does not rely on a privileged teacher policy, nor expert demonstrations during training. \methodname{} consists of a frozen ViT encoder, learned attention pooling, an encoder-only transformer actor; it is trained entirely in simulation with asymmetric actor-critic RL~\cite{pinto2018asymmetric} (PPO) before being deployed directly in the real world.
\subsection{Algorithmic Decisions for RL Training}
\label{sec:rl_details}%
We directly train visual policies with distributed Proximal Policy Optimization~\cite{schulman2017proximal} (PPO) using an asymmetric critic $v_\psi(s_t^{\mathrm{priv}})$ that accesses privileged simulator state during training (Fig.~\ref{fig:method}, right). The actor is a three-layer encoder-only transformer with hidden dimension 64 that uses bidirectional self-attention. Similar to Dosovitskiy et al.~\cite{dosovitskiy2020image}, we add learnable position embeddings to tokens by type. Self-attention jointly integrates visual and proprioceptive tokens through a common representation that can accommodate different visual encoders, camera configurations, and per-view token budgets. We parallelize rollouts and synchronize PPO updates across GPU workers, using Adam~\cite{kingma2014adam} with a constant base learning rate of $3\times10^{-4}$.

\subsection{Frozen Visual Features}
\begin{figure}[!t]
    \centering
    \includegraphics[width=\columnwidth]{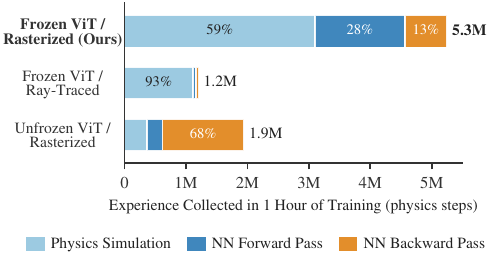}
    \caption{\textbf{Frozen ViT features and rasterization substantially increase training throughput.} Visual RL breakdown per training hour on \texttt{Peg-Insertion}. Bar length gives physics steps collected per hour; segments show the fraction of timed training spent in simulation (including rendering), the neural-network forward pass, and backpropagation. Fractions are re-normalized over these three components; evaluation and un-timed overhead are under 2\%. Rows compare rasterization with ray tracing and a frozen with an unfrozen DINOv3 ViT-S/16 encoder. Ray tracing makes simulation dominate; unfreezing the encoder makes backpropagation dominate.}
    \label{fig:compute_hour_ppo}
\end{figure}
\label{sec:frozen_visual_feats}
A frozen encoder $f_{\phi}$ produces patch embeddings
\begin{equation}
    X_t =
    \operatorname{sg}\!\left[f_{\phi}\!\left(g(I_t)\right)\right]
    \in\mathbb{R}^{N\times d_e},
    \qquad \nabla_\phi\mathcal{L}_{\mathrm{RL}}=\mathbf{0},
    \label{eq:frozen_features}
\end{equation}where $\operatorname{sg}$ denotes the stop-gradient operator, and $g$ denote the desired crop, resize, and encoder normalization. Using frozen patch embeddings allows us to avoid back-propagating through an expressive visual encoder, which is a primary bottleneck for visual RL training.
Freezing the visual encoder also amortizes the forward-pass cost across the $K$ PPO epochs within each policy improvement step. Computing patch embeddings once and reusing them across epochs reduces encoder forward-pass compute by a factor of $K$, leading to a further increase in training throughput. Together, these advantages yield a substantial increase in the amount of experience \methodname{} collects in an hour of training (Fig.~\ref{fig:compute_hour_ppo}, top), compared with unfreezing the encoder (Fig.~\ref{fig:compute_hour_ppo}, bottom).

Furthermore, freezing preserves the visual representations learned through large-scale pretraining on natural images, avoiding encoder updates that could specialize these features to the appearance of simulated images. We find that these representations support zero-shot sim-to-real transfer even when policies are trained with fast rasterized images. This allows us to avoid the substantial cost of high-fidelity ray-traced rendering, which also limits the amount of experience that can be collected in an hour of training (Fig.~\ref{fig:compute_hour_ppo}, middle).

We find that frozen $\texttt{[CLS]}$ features alone are insufficient for precision manipulation, consistent with recent
  findings~\cite{zhou2026patch}. Processing all DINOv3 patch embeddings is computationally expensive. Therefore, we employ a lightweight multihead attention pooling (MAP) head~\cite{jaegle2021perceiver,zhai2022scaling, alayrac2022flamingo}, whose learnable queries
  cross-attend to the patch embeddings to produce a compact sequence
  of visual tokens for the transformer policy. This significantly reduces the cost of downstream self-attention which scales quadratically with the number of input tokens.
\section{Experimental Setup}
\begin{figure}[!t]
    \centering
    \includegraphics[width=\columnwidth]{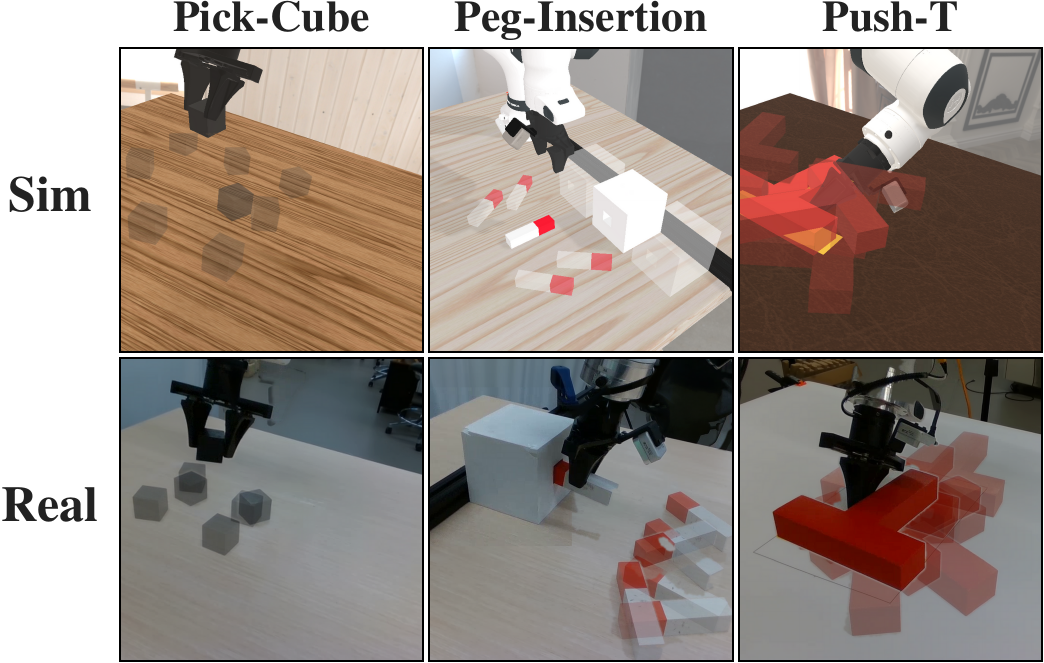}
    \caption{\textbf{Tasks in simulation and the real world.} Initial-state views for Pick-Cube, Peg-Insertion, and Push-T in simulation (top) and the real world (bottom). Translucent objects in the simulation row illustrate different initial poses for the objects. The goal state in the realworld is shown as well.}
    \label{fig:task_teaser}
\end{figure}
\begin{figure*}[!t]
    \centering
    \includegraphics[width=\textwidth]{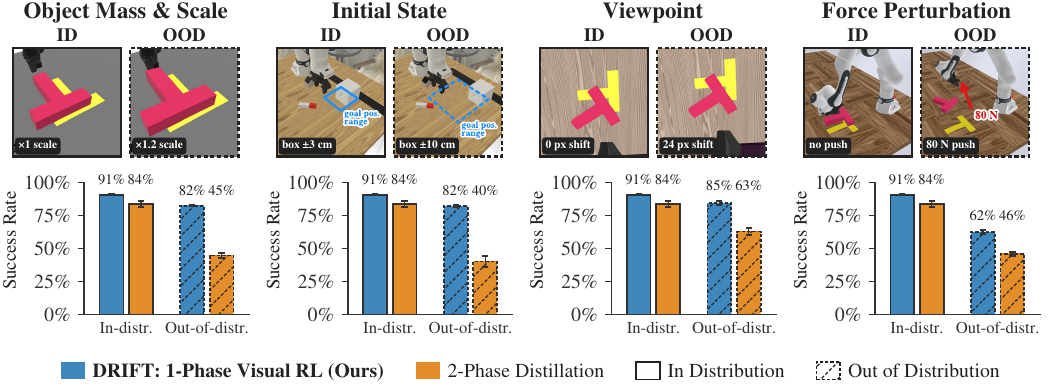}
    \caption{\textbf{One-phase visual RL retains higher success under distribution shifts.} In-distribution (In-distr.) and selected out-of-distribution (Out-of-distr., hatched) success for 1-Phase Visual RL and 2-Phase Distillation (DAgger with a frozen encoder). Object mass and scale use $4\times$ mass for both tasks and $1.2\times$ tee and goal scale for \texttt{Push-T}. Initial-state shifts use a 2\,cm goal-position ring for \texttt{Push-T} (training jitter $\pm1$\,cm), a 10\,cm box-position ring for \texttt{Peg-Insertion} (training range $\pm5$\,cm), and peg-start offsets in $(1,1.25]$ times the training half-range; rings use the maximum absolute axis offset. Viewpoint shifts use a 24\,px image translation for \texttt{Push-T} and $0.85\times$ camera field of view for \texttt{Peg-Insertion}. Force perturbations are \texttt{Push-T} only: 0.2\,s pushes of 80\,N on the gripper or 3\,N on the tee. Success is episode-weighted within each selected sweep, then averaged equally across sweeps within each task and across tasks where available. Bars show means over three paired training seeds; error bars show $\pm1$ standard error of the mean. Top-row images are illustrative examples.}
    \label{fig:generalization_summary}
\end{figure*}
\begin{figure}[!t]
    \centering
    \includegraphics[width=\columnwidth]{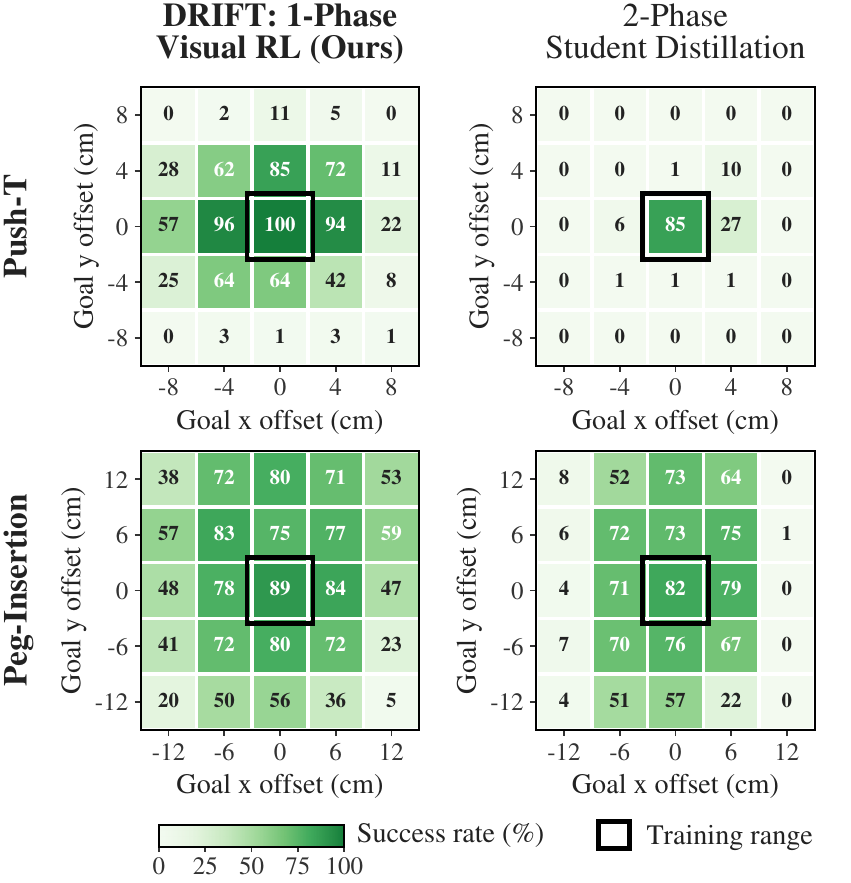}
    \caption{\textbf{One-phase RL generalizes over a wide range of goals.} Simulation success for \texttt{Push-T} goal positions (top) and \texttt{Peg-Insertion} box positions (bottom), comparing 1-Phase Visual RL (left) with 2-Phase Student Distillation using DAgger with a frozen encoder (right). Each panel subsamples a $5\!\times\!5$ grid of 2\,cm evaluation cells at the labeled offsets; numbers show mean success (\%) over three training seeds and 100 episodes per seed and cell. Black squares mark the displayed offsets within the training range ($\pm$1\,cm for \texttt{Push-T}; $\pm$5\,cm for \texttt{Peg-Insertion}), represented by the center cell in each panel.}
    \label{fig:init_state_heatmaps}
\end{figure}

\subsection{Task Design}
\label{sec:tasks}
We select three tasks (Fig.~\ref{fig:task_teaser}) spanning simple grasping, non-prehensile pushing, and precision insertion: 1) \texttt{Pick-Cube}, grasping and lifting a cube; 2) \texttt{Push-T}, aligning a T-shaped block with a visible goal through non-prehensile pushing; and 3) \texttt{Peg-Insertion}, inserting a square peg into a hole with $2$mm clearance through precise, contact-rich interaction. Object and goal initial poses are randomized across tasks.

\subsection{Training Details}
We use ManiSkill~\cite{tao2025tongzhou} with a shared vectorized simulation environment across tasks.
The action space is $a=[\delta q,g]\in\mathbb{R}^{8}$, specifying relative arm joint-position targets and the desired gripper position. Policies observe RGB images and proprioception, $o=(I,p)$, where $p\in\mathbb{R}^{15}$ contains joint positions, joint velocities, and measured gripper position; \texttt{Push-T} omits gripper position, yielding $p\in\mathbb{R}^{14}$. Simulated cameras and robot actuator dynamics are calibrated against the real system nominal parameters, as detailed in sections~\ref{sec:camera_cal} and~\ref{sec:sysid} in the appendix. To promote sim-to-real transfer, we randomize object friction and appearance, table textures, lighting through sampled HDR environment maps, and camera extrinsics around the calibrated values. We additionally replace image background pixels using simulator segmentation masks with approximately $27$k natural images.

\subsection{Evaluation Protocol}
\label{sec:eval_protocol}
We deploy policies at $20$Hz on a Franka Research 3 (FR3) arm, using RGB images from a wrist camera and proprioception observations. We report \textit{success rate} and \textit{throughput}, defined as successful task completions per unit of rollout time, counting successful episodes until first success and failed episodes for the full horizon. For each method and task, we periodically evaluate training checkpoints over $1{,}024$ simulation episodes and retain the best checkpoint from each of three seeds.
We select among these checkpoints using five real-world rollouts each, then report the selected checkpoint’s performance over $20$ or $30$ rollouts.

\section{Experimental Results}
We answer the following questions:
\begin{enumerate}[leftmargin=*, label=\Alph*., nosep]
    \item How does \methodname{} compare with two-phase distillation under in-distribution (ID) and out-of-distribution (OOD) settings in simulation, given matched training compute budgets?
    \item Does \methodname{} transfer more effectively to the real world than two-phase distillation?
    \item Which visual encoder yields the best sim-to-real success?
\end{enumerate}
\subsection{In-Distribution and Out-of-Distribution Sim Evaluations} \label{sec:sim-id-ood}

\textbf{Setup.}
We compare \methodname{} against two-phase distillation under matched architectures and compute budgets. Both methods use the same frozen DINO visual encoder and are trained for 150 NVIDIA L40S equivalent GPU hours. We evaluate their robustness to distribution shift on the \texttt{Peg-Insertion} and \texttt{Push-T} tasks by measuring performance under both in-distribution (ID) and out-of-distribution (OOD) conditions. In Fig.~\ref{fig:generalization_summary}, we evaluate four sources of distribution shift. We evaluate four distribution shifts: (i) \texttt{Object Properties}, varying mass up to $4\times$ and scale up to $1.2\times$; (ii) \texttt{Initial-State Distribution}, expanding goal-position ranges from $\pm1$ to $\pm2$cm for \texttt{Push-T} and $\pm5$ to $\pm10$cm for \texttt{Peg-Insertion}; (iii) \texttt{Viewpoint}, applying 24-pixel translations and $0.85\times$ field-of-view scaling; and (iv) \texttt{Force Perturbations}, applying $80$N random forces to the gripper. Fig.~\ref{fig:init_state_heatmaps} further evaluates generalization over a wider range of initial goal positions.

\textbf{Results.}
Figure~\ref{fig:generalization_summary} summarizes the aggregate results. Under in-distribution (ID) conditions, the two methods achieve broadly similar performance, with \methodname{} attaining an 8.6\% relative improvement in average success rate over two-phase distillation. Under OOD shifts, however, two-phase distillation loses approximately 42\% of its ID success rate, compared with only 14\% for \methodname{}. Consequently, \methodname{}'s relative advantage grows to 61\% in OOD settings, demonstrating substantially stronger generalization beyond the training distribution.

We next examine goal-position generalization in greater detail in Fig.~\ref{fig:init_state_heatmaps}. Within the range of goal poses encountered during training, the two methods again perform comparably. Outside this training distribution, however, the performance of two-phase distillation degrades much more rapidly, while \methodname{} retains high success over a broader region.

Together, these results suggest that directly optimizing the visual policy with RL produces policies that are substantially more robust to distribution shift than those obtained through two-phase privileged-policy distillation. This improved OOD robustness may also be important for sim-to-real transfer, where discrepancies between simulation and the real world inevitably induce distribution shift at deployment.
\begin{table*}[t!]
\centering

\footnotesize
\renewcommand{\thetable}{\arabic{table}}
\makeatletter
\renewcommand{\fnum@table}{Table~\thetable}
\long\def\@makecaption#1#2{%
    \vskip 5pt%
    \parbox[t]{\hsize}{\footnotesize\noindent\textbf{#1:} #2\par}%
}
\makeatother

{\normalfont\footnotesize Real-world transfer across manipulation tasks.\par}
\vspace{5pt}
\definecolor{transferblue}{HTML}{4C83D3}
\setlength{\tabcolsep}{3pt}
\renewcommand{\arraystretch}{1.3}

\newcommand{\transferpoint}[1]{\makebox[3.4em][r]{$#1$}}
\newcommand{\transferbest}[1]{\makebox[3.4em][r]{$\mathbf{#1}$}}
\newcommand{\transfermethod}{%
    \textcolor{transferblue!65!black}{%
        \textbf{\methodname{}: 1-phase visual RL (ours)}}%
}

\begin{tabular*}{\textwidth}{@{\extracolsep{\fill}}llccccc@{}}
\toprule

& &
\multicolumn{2}{c}{\textbf{Real-world performance}} &
\multicolumn{3}{c}{\textbf{Simulation-to-real transfer}}
\\
\cmidrule(lr){3-4}
\cmidrule(lr){5-7}

\textbf{Task} &
\textbf{Method} &
\shortstack{\textbf{Success} $\uparrow$\\
            \textcolor{black!65}{(\%)}} &
\shortstack{\textbf{Throughput} $\uparrow$\\
            \textcolor{black!65}{(successes/min)}} &
\shortstack{\textbf{Success in sim} $\uparrow$\\
            \textcolor{black!65}{(\%)}} &
\shortstack{\textbf{Sim2Real Drop} $\downarrow$\\
            \textcolor{black!65}{(percentage points)}} &
\shortstack{\textbf{Real}\\
            \textbf{trials}}
\\
\midrule

\textbf{Peg insertion}
& \transfermethod
& \transferbest{66.7}
& \transferbest{4.31}
& \transferbest{82.3}
& \transferbest{15.7}
& \textbf{30}
\\

& 2-phase distillation (w/ frozen ViT)
& \transferpoint{23.3}
& \transferpoint{1.13}
& \transferpoint{78.5}
& \transferpoint{55.2}
& 30
\\

& 2-phase distillation (w/ unfrozen ViT)
& \transferpoint{6.7}
& \transferpoint{0.28}
& \transferpoint{74.6}
& \transferpoint{67.9}
& 30
\\

\midrule

\textbf{Pick-Cube}
& \transfermethod
& \transferbest{100.0}
& \transferbest{28.47}
& \transferbest{99.7}
& \transferbest{-0.3}
& \textbf{20}
\\

& 2-phase distillation (w/ frozen ViT)
& \transferpoint{100.0}
& \transferpoint{28.74}
& \transferpoint{96.4}
& \transferpoint{-3.6}
& 20
\\

\midrule

\textbf{Push-T}
& \transfermethod
& \transferbest{76.7}
& \transferbest{5.54}
& \transferbest{99.3}
& \transferbest{22.6}
& \textbf{30}
\\

& 2-phase distillation (w/ frozen ViT)
& \transferpoint{30.0}
& \transferpoint{1.46}
& \transferpoint{93.6}
& \transferpoint{63.6}
& 30
\\

& 2-phase distillation (w/ unfrozen ViT)
& \transferpoint{16.7}
& \transferpoint{0.75}
& \transferpoint{92.1}
& \transferpoint{75.4}
& 30
\\

\bottomrule
\end{tabular*}

\caption[Real-world transfer across manipulation tasks.]{%
\textbf{Zero-Shot Sim-to-Real Transfer.} \methodname{} achieves slightly higher success in simulation, but substantially better real-world performance across all three tasks, resulting in markedly smaller sim-to-real degradation than two-phase distillation. For each method, we report the seed with the highest real-world success. Simulation success is evaluated over 1,024 episodes per checkpoint, and throughput measures successful task completions per minute of rollout time.
}
\label{tab:real_transfer_summary}

\end{table*}
\begin{figure*}[!t]
    \centering
    \includegraphics[width=\textwidth]{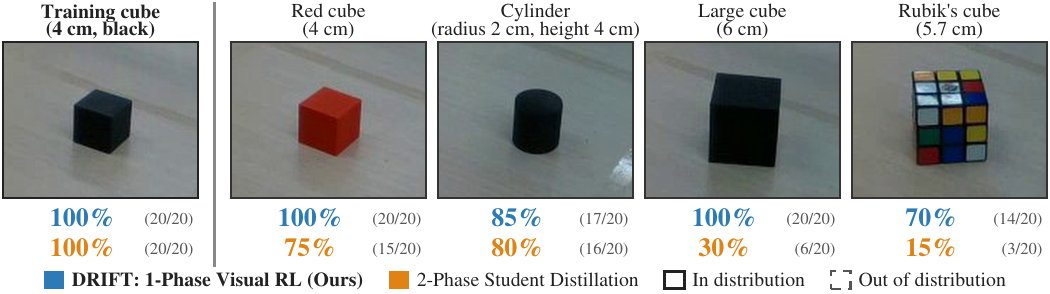}
    \caption{\textbf{One-phase visual RL transfers better to unseen real-world objects than two-phase distillation.} Policies are trained in simulation on a single black 4,cm cube and evaluated zero-shot on four unseen objects over 20 real-world trials each. Both methods achieve 100\% success on the training object, while \methodname{} retains substantially higher success on objects with novel appearance and geometry.}
    \label{fig:pickcube_real_ood}
\end{figure*}
\begin{figure}[!t]
    \centering
    \includegraphics[width=\columnwidth]{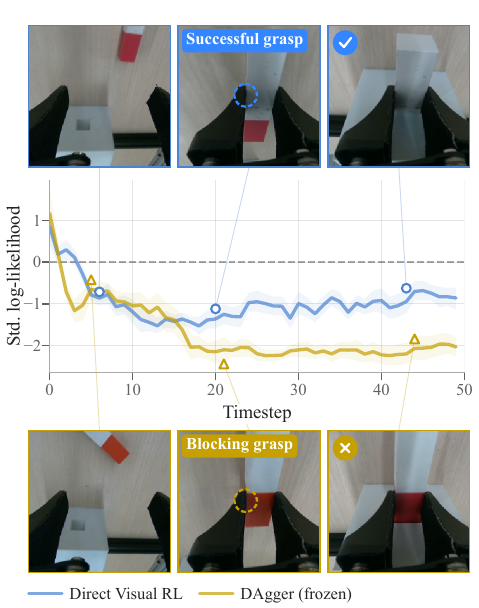}
    \caption{\textbf{\methodname{} real-world latents remain more consistent with simulation.} Mean standardized latent log-likelihood over 30 real-world episodes per policy. Scores are standardized using each policy's mean and standard deviation across completed held-out simulation rollouts. Shading denotes pointwise 95\% bootstrap confidence intervals over real-world episodes. \methodname{} (blue) exhibits a smaller standardized likelihood deficit than two-phase distillation (gold), consistent with its real-world latents remaining closer to its simulation distribution throughout rollout.}
    \label{fig:latent_typicality}
\end{figure}

\subsection{Zero-Shot Sim-to-Real Transfer Results}
\label{sec:real_world_transfer}
\textbf{Setup.}
We evaluate zero-shot sim-to-real transfer across the three tasks in Section~\ref{sec:tasks}, deploying policies trained entirely in simulation without real-world data or real-world gradient updates. We compare \methodname{} against two-phase distillation with either a frozen or unfrozen visual encoder, using matched architectures and compute budgets. For each method, we select checkpoints according to Section~\ref{sec:eval_protocol} and report real-world success rate and throughput alongside the corresponding simulation success in Table~\ref{tab:real_transfer_summary}.

\textbf{Results.}
Despite only modest differences in simulation, various methods exhibit substantially different real-world performance. \methodname{} achieves slightly higher simulation success across all three tasks, but its advantage becomes much larger after transfer. On \texttt{Peg-Insertion}, \methodname{} achieves $66.7\%$ real-world success compared with $23.3\%$ for two-phase distillation with a frozen ViT, while reducing the sim-to-real drop from $55.2$ to $15.7$ percentage points. Similarly, on \texttt{Push-T}, \methodname{} reaches $76.7\%$ real-world success compared with $30.0\%$ for the frozen-ViT baseline, corresponding to sim-to-real drops of $22.6$ and $63.6$ percentage points, respectively. Unfreezing the visual encoder during distillation further degrades transfer on both tasks. This demonstrates that similar simulation performance can conceal substantial differences in robustness to deployment shifts, with direct visual RL transferring more reliably than two-phase distillation.

On the simpler \texttt{Pick-Cube} task, both approaches transfer successfully, achieving $100\%$ real-world success. However, this parity does not persist when evaluating the policy on unseen objects (Fig.~\ref{fig:pickcube_real_ood}). Across four unseen objects with $20$ real-world trials each, \methodname{} achieves an average success rate of $88.8\%$, compared with $50.0\%$ for two-phase distillation. Thus, even on this simpler task, direct RL exhibits stronger robustness to changes in object appearance and geometry.

These results suggest that errors induced by the sim-to-real gap more readily cause failure in tasks requiring precise or extended interaction. Even on simpler tasks, increasing the deployment shift through unseen objects reveals the same robustness advantage for direct RL.

To examine these transfer differences, we compare real-world policy latents with their respective simulation rollout distributions in Fig.~\ref{fig:latent_typicality}, motivated by likelihood-based typicality analysis~\cite{nalisnick2019detecting}. We fit a separate Gaussian mixture model (GMM) to each policy's transformer latents collected during simulation rollouts. We evaluate the log-likelihood of real-world latents under the corresponding estimated simulation density and standardize log-likelihoods using each policy's mean and standard deviation across completed held-out simulation rollouts. Under the same real-world evaluation conditions, \methodname{} exhibits a smaller standardized likelihood deficit than two-phase distillation during execution. This is consistent with direct RL learning more robust behavior, with its real-world latent representations remaining more typical of its simulated rollout distribution.

Real-world deployment introduces shifts in visual observations and physical dynamics that persist even after real-to-sim alignment. These shifts stress the generalization capabilities of the policy, in a similar way as the controlled OOD conditions in simulation (Section~\ref{sec:sim-id-ood}). \methodname{} outperforms two-phase distillation by similar relative margins in both settings: approximately $60\%$ higher success under OOD simulation conditions and $58.8\%$ higher mean real-world success. This suggests that its stronger sim-to-real transfer stems in part from the same robustness to distribution shift observed in simulation.

\begin{figure}[!t]
    \centering
    \includegraphics[width=\columnwidth]{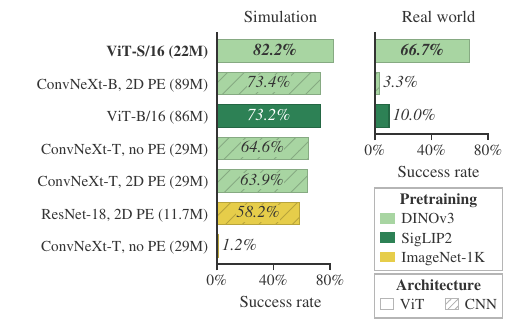}
    \caption{\textbf{Not all frozen visual representations are equal for transfer.} Frozen visual encoder ablation on \texttt{Peg-Insertion} for seven selected encoders in simulation (left), and for the best three encoders in the real world (right). DINOv3-S/16 outperforms all other representations by a small margin in simulation, and by a significant margin in the real world.}
    \label{fig:frozen_encoder_sim_real}
\end{figure}
\subsection{Which Vision Encoder Transfers Best to the Real-world?}
\label{sec:encoder_ablation}
\textbf{Setup.}
We compare seven visual encoder configurations on \texttt{Peg-Insertion}, spanning ViT and CNN backbones with DINOv3, SigLIP2, and ImageNet-1K pretraining. All encoders remain frozen during visual RL training. We report the simulation success rate across $512$ evaluation episodes per encoder. We also evaluate sim-to-real transferability with $30$ real world trials for the three best performing encoders (highest simulation success rate).

\textbf{Results.}
Among the configurations shown, DINOv3 ViT-S/16 achieves the highest simulation success at $82.2\%$ and the strongest real-world performance at $66.7\%$ (Fig. \ref{fig:frozen_encoder_sim_real}). DINOv3 ConvNeXt-B with fixed sinuisoidal positional encodings added to the reshaped feature map and SigLIP2 ViT-B/16 also attain substantial simulation success, at $73.4\%$ and $73.2\%$, respectively. However, their real-world success falls to just $3.3\%$ and $10.0\%$. Despite their larger model sizes, these encoders transfer substantially worse than the $22$M-parameter DINOv3 ViT-S/16. These results motivate the choice of DINOv3 ViT-S/16 for this setting, and suggest that strong simulation performance alone does not ensure effective real-world transfer.
\section{Conclusion and remarks}

We present DRIFT, a simple recipe for efficient one-phase visual RL that combines frozen pretrained vision encoders, learned attention pooling, and fast rasterized simulation. These choices address the computational burden of visual learning while preserving the spatial detail that precise manipulation demands and the robustness that real-world transfer requires. By making such representations inexpensive to use, we free more of the GPU budget for experience collection, making it practical to learn transferable manipulation policies directly from pixels.

Beyond training efficiency, DRIFT transfers substantially better than the standard two-phase distillation pipeline, achieving a 58.8\% relative improvement in average real-world success across three manipulation tasks under matched architectures, observations, and compute budgets. Although the two approaches attain broadly similar performance within the simulation training distribution, DRIFT retains higher success under controlled distribution shifts and on unseen real-world objects. These results suggest that directly optimizing the visual policy with RL learns more robust behavior, helping explain why its advantage becomes substantially larger when policies encounter the discrepancies between simulation and the real world.

\FloatBarrier

\onecolumn
\appendix

\makeatletter
\newcommand{\appendixtablestyle}{%
    \footnotesize
    \renewcommand{\thetable}{\arabic{table}}%
    \renewcommand{\fnum@table}{Table~\thetable}%
    \long\def\@makecaption##1##2{%
        \vskip 5pt%
        \parbox[t]{\hsize}{\footnotesize\noindent\textbf{##1:} ##2\par}%
    }%
}
\makeatother

\subsection{One-Phase Policy Optimization}
\label{sec:ppo_details}
We learn visual policies with proximal policy optimization, Schulman et
al.~\cite{schulman2017proximal}, estimating advantages with generalized advantage
estimation~\cite{schulman2015high}. After each rollout we run up to $K$ epochs of
minibatch updates, stopping early once
\mbox{$\hat{D}_{\mathrm{KL}}(\pi_{\theta_{\mathrm{old}}}\,\|\,\pi_\theta)>d_{\mathrm{targ}}$}.
The asymmetric critic is trained jointly with an unclipped squared-error value loss.

\begin{table}[!ht]
\centering
\appendixtablestyle
\renewcommand{\arraystretch}{1.15}
\begin{tabular}{@{}p{0.5\textwidth}>{\centering\arraybackslash}p{\dimexpr0.5\textwidth-2\tabcolsep\relax}@{}}
\toprule
\textbf{Parameter} & \textbf{Value} \\
\midrule
Clipping parameter $\epsilon$ & 0.2 \\
Value loss coefficient & 0.5 \\
Advantage normalization & True \\
Target KL divergence $d_{\mathrm{targ}}$ & 0.1 \\
PPO epochs per update $K$ & 4 \\
Learning rate & $3\cdot10^{-4}$ \\
Max gradient norm & 0.5 \\
Parallel environments & 1024 \\
Rollout horizon (steps) & 32 \\
Minibatch size & 1024 \\
\bottomrule
\end{tabular}
\caption{\textbf{One-phase visual RL training hyperparameters.} Settings used for all
visual RL policies. \texttt{Pick-Cube} policies were trained on a single NVIDIA L40S
GPU; all other policies were trained with distributed training using 4 L40S GPUs.}
\label{tab:ppo_hparams}
\end{table}

\begin{table}[!ht]
\centering
\appendixtablestyle
\renewcommand{\arraystretch}{1.15}
\begin{tabular}{@{}p{0.5\textwidth}>{\centering\arraybackslash}p{\dimexpr0.5\textwidth-2\tabcolsep\relax}@{}}
\toprule
\textbf{Parameter} & \textbf{Value} \\
\midrule
Image resolution & (224, 224, 3) \\
Visual encoder & DINOv3 ViT-S/16 \\
Visual encoder weights & Frozen \\
Visual encoder output & (196, 384) \\
Attention pooling input projection & (384, 64) \\
Attention pooling heads & 8 \\
Attention pooling feedforward dimension & 256 \\
Proprioception MLP layers & 2 \\
Proprioception MLP width & 64 \\
Actor activations & GELU \\
Transformer type & Encoder-only \\
Transformer blocks & 3 \\
Transformer embedding dimension & 64 \\
Transformer attention heads & 8 \\
Transformer feedforward dimension & 256 \\
Transformer normalization & LayerNorm (pre-norm) \\
Transformer dropout & 0 \\
Actor readout & [CLS] \\
Readout projections & Linear \\
Critic MLP hidden dimension & 256 \\
Critic MLP hidden layers & 3 \\
\bottomrule
\end{tabular}
\caption{\textbf{One-phase visual RL network architecture.} Policies trained with
one-phase RL and with two-phase distillation use the same visual actor architecture.
Self-attention is bidirectional for all transformers.}
\label{tab:architecture}
\end{table}

\subsection{Camera Calibration}
\label{sec:camera_cal}
We initialize simulated cameras from calibrated real-world intrinsics and
camera poses estimated relative to the robot through hand--eye calibration.
At deployment, images are center-cropped to a square and resized to the policy
resolution. We incorporate this image-coordinate transformation into the
nominal simulated intrinsics as $K_{\mathrm{sim}} = A K_{\mathrm{real}}$, where
$A$ is the crop-and-resize transform.\footnote{With focal lengths $\mathbf f=(f_x,f_y)$,
principal point $\mathbf c=(c_x,c_y)$, crop offset $\mathbf o$, and resize
factor $s$, $A$ gives $\mathbf f'=s\mathbf f$ and
$\mathbf c'=s(\mathbf c-\mathbf o)$ for $K_{\text{sim}}$.}
Together with matched extrinsics, this aligns viewing rays and projected object scales between
simulation and preprocessed real images under the pinhole camera model.

\subsection{Actuator System Identification}
\label{sec:sysid}
We record joint positions and velocities under sinusoidal chirp commands on
the real robot and replay the same position targets in simulation. Using
CMA-ES~\cite{6790790}, we jointly fit PD gains, armature, and Coulomb friction
coefficients for the seven arm joints by minimizing the sum of position and
velocity mean-squared errors between simulated and recorded trajectories,
averaged over time and joints. Each rollout starts from the recorded initial
joint positions with zero velocity; torque and torque-rate limits remain fixed
during fitting. To model the FR3's torque-rate constraint $|\dot{\tau}|\leq\dot{\tau}_{\max}$,
we rate-limit the PD torque command at each physics step $k$. For each joint,
with step size $\Delta t$ and $\delta=\dot{\tau}_{\max}\Delta t$, we compute:
\begin{align}
\tau_k^{\mathrm{PD}} &= K_p(q_{k}^d-q_k)-K_d\dot q_k,\\
\tau_k &= \tau_{k-1}+\operatorname{clip}
  (\tau_k^{\mathrm{PD}}-\tau_{k-1},-\delta,\delta),\\
\tilde q_{k}^{d} &= q_k+ K_p^{-1}\left(\tau_k+K_d\dot q_k\right).
\end{align}
Here $\tau_{k-1}$ is the previous rate-limited torque command. To enforce torque-rate constraints, the
adjusted position targets $\tilde q_k^{\mathrm{des}}$ are passed to the simulator's implicit
joint drive, using $\dot{\tau}_{\max}=990\,\mathrm{N\,m\,s^{-1}}$.
The physics engine enforces the joint torque limits.

\end{document}